\documentclass[10pt,twocolumn,letterpaper]{article}

\usepackage[pagenumbers]{cvpr} 

\usepackage{graphicx}
\usepackage{amsmath}
\usepackage{amssymb}
\usepackage{booktabs}
\usepackage[accsupp]{axessibility}

\usepackage[pagebackref,breaklinks,colorlinks,bookmarksdepth=3,bookmarksopen=true,bookmarksopenlevel=2,bookmarksnumbered=true]{hyperref}

\usepackage[capitalize]{cleveref}
\crefname{section}{Sec.}{Secs.}
\Crefname{section}{Section}{Sections}
\Crefname{table}{Table}{Tables}
\crefname{table}{Tab.}{Tabs.}

\usepackage{adjustbox}

\def\cvprPaperID{1389} 
\def\confName{CVPR}
\def\confYear{2023}

\newcommand{\parsection}[1]{\noindent\textbf{#1.}~}

\begin{document}

\title{HandNeRF: Neural Radiance Fields for Animatable Interacting Hands}


\author{
Zhiyang Guo$^{1}$ \and
Wengang Zhou$^{1,2}$\footnotemark[1] \and
Min Wang$^{2}$ \and
Li Li$^{1}$ \and
Houqiang Li$^{1,2}$\footnotemark[1] \and
$^{1}$CAS Key Laboratory of Technology in GIPAS, EEIS Department,\\
University of Science and Technology of China \\
$^{2}$Institute of Artificial Intelligence,
Hefei Comprehensive National Science Center \\
{\tt\small guozhiyang@mail.ustc.edu.cn, \{zhwg, lil1, lihq\}@ustc.edu.cn, wangmin@iai.ustc.edu.cn}
\vspace{-2mm}
}


\twocolumn[{
\maketitle
\vspace{-10mm}
\begin{center}
  \centering
  \includegraphics[width=\linewidth]{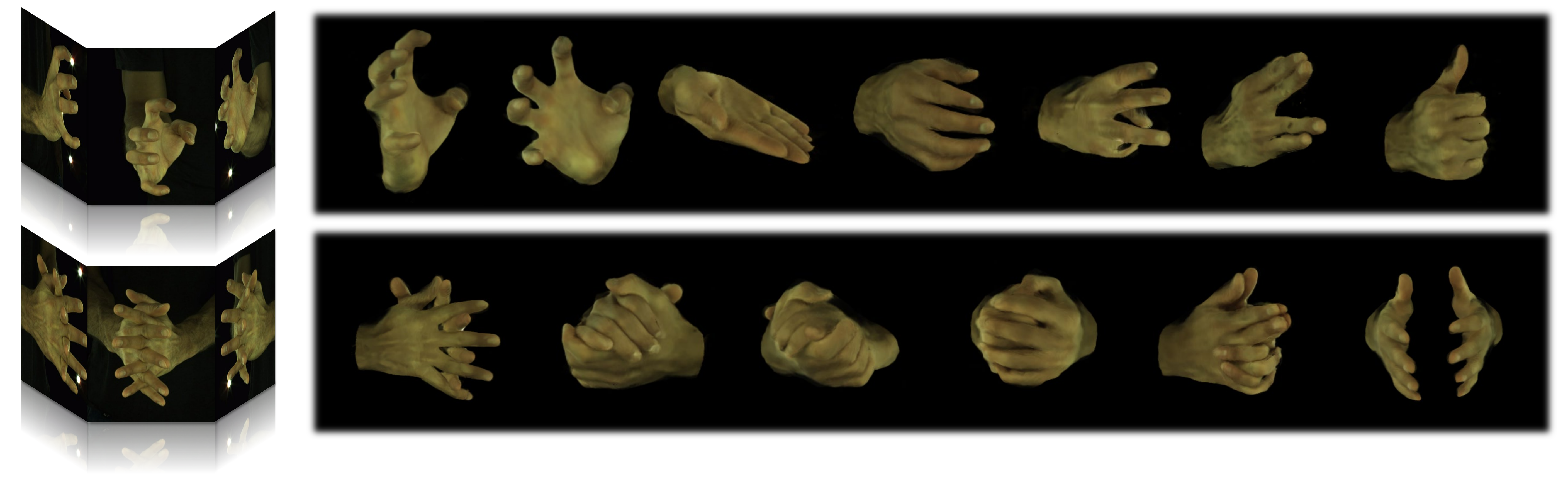}
\end{center}
\vspace{-8mm}
\captionof{figure}{Given a set of multi-view images capturing a pose sequence of a single hand or two interacting hands (left), HandNeRF models the scene in a unified manner with neural radiance fields, enabling rendering of novel hand poses from arbitrary viewing directions (right).}
\label{fig:teaser}
\vspace{6mm}
}]

{
  \renewcommand{\thefootnote}%
    {\fnsymbol{footnote}}
  \footnotetext[1]{Corresponding Authors.}
}

\begin{abstract}
  We propose a novel framework to reconstruct accurate appearance and geometry with neural radiance fields (NeRF) for interacting hands, enabling the rendering of photo-realistic images and videos for gesture animation from arbitrary views.
  Given multi-view images of a single hand or interacting hands, an off-the-shelf skeleton estimator is first employed to parameterize the hand poses. Then we design a pose-driven deformation field to establish correspondence from those different poses to a shared canonical space, where a pose-disentangled NeRF for one hand is optimized. Such unified modeling efficiently complements the geometry and texture cues in rarely-observed areas for both hands.
  Meanwhile, we further leverage the pose priors to generate pseudo depth maps as guidance for occlusion-aware density learning. Moreover, a neural feature distillation method is proposed to achieve cross-domain alignment for color optimization.
  We conduct extensive experiments to verify the merits of our proposed HandNeRF and report a series of state-of-the-art results both qualitatively and quantitatively on the large-scale InterHand2.6M dataset.
\end{abstract}


\section{Introduction}
\label{sec:intro}
As a dexterous tool to interact with the physical world and convey rich semantic information, the modeling and reconstruction of human hands have attracted substantial attention from the research community.
Typically, the synthesis of realistic hand images or videos with different postures in motion has a wide range of applications, \eg, human-computer interaction, sign language production, virtual and augmented reality technologies such as telepresence, \etc.

Classic hand-modeling works are mainly built upon parameterized mesh models such as MANO~\cite{mano}. They fit the geometry of hands to polygon meshes manipulated by shape and pose parameters, and then complete coloring via texture mapping. Despite being widely adopted, those models have the following limitations. On the one hand, high-frequency details are hard to present on low-resolution meshes, hindering the production of photo-realistic images. On the other hand, no special design is developed for interacting hands, which is a non-trivial scenario involving complex postures with self-occlusion.

To address the above issues and push the boundary of realistic human hand modeling, motivated by the recent success of NeRF\cite{nerf} in modeling human body~\cite{neuralbody,aninerf,neuman}, we propose \textbf{HandNeRF}, a novel framework that unifiedly models the geometry and texture of animatable interacting hands with neural radiance fields (NeRF).
Specifically, a pose-conditioned deformation field is introduced to warp the sampled observing ray into a canonical space, guided by the prior-based blend skinning transformation and a learnable error-correction network dealing with non-rigid deformations. The different input postures are thereby mapped to a common mean pose, where a canonical NeRF is competent at modeling. Thanks to the continuous implicit representation of NeRF and the multi-view-consistent volume rendering, we are able to produce high-fidelity images of posed hands from arbitrary viewing directions. This can not only be applied in the synthesis of free-viewpoint videos, but also help to perform data augmentation for multi-view detection and recognition tasks in computer vision, \eg, sign language recognition.

Meanwhile, modeling one single hand is nowhere near enough from an application perspective. The semantics expressed by single-hand movements is quite limited. Many practical scenarios such as sign language conversations require complex interacting postures of both hands.
However, handling interaction scenarios is far from trivial and still lacks exploration. Interacting hands exhibit fine-grained texture in small areas, while incompleteness of visible texture permeates the image samples due to self-occlusion and limited viewpoints. To this end, we extend the aforementioned model into a unified framework for both hands. By introducing the hand mapping and ray composition strategy into the pose-deformable NeRF, we make it possible to naturally handle interaction contacts and complement the geometry and texture in rarely-observed areas for both hands.
Note that with such a design, HandNeRF is compatible with both single hand and two interacting hands.

Moreover, to ensure a correct depth relationship when rendering the hand interactions, we re-exploit the human priors and propose a low-cost depth supervision for occlusion-robust density optimization. Such strong constraint guides the model to extract accurate geometry from sparse-view training samples.
Additionally, a neural feature distillation branch is designed to achieve feature alignment between a pre-trained 2D teacher and the 3D color field. By implicitly leveraging spatial contextual cues for color learning, this cross-domain distillation effectively alleviates the artifacts on the target shape and further improves the quality of the learned texture.

Our main contributions are summarized as follows:
\vspace{-2mm}

\begin{itemize}
    \item To the best of our knowledge, we are the first to develop a unified framework to model photo-realistic interacting hands with deformable neural radiance fields.
    \vspace{-5mm}

    \item We propose several elaborate strategies, including the depth-guided density optimization and the neural feature distillation, in order to effectively address practical challenges in interacting hands training and ensure high-fidelity results for novel view/pose synthesis.
    \vspace{-5mm}

    \item Extensive experiments on the large-scale dataset InterHand2.6M~\cite{interhand} show that our HandNeRF outperforms the baselines both qualitatively and quantitatively.
\end{itemize}

\begin{figure*}[t]
  \centering
  \includegraphics[width=1.0\linewidth]{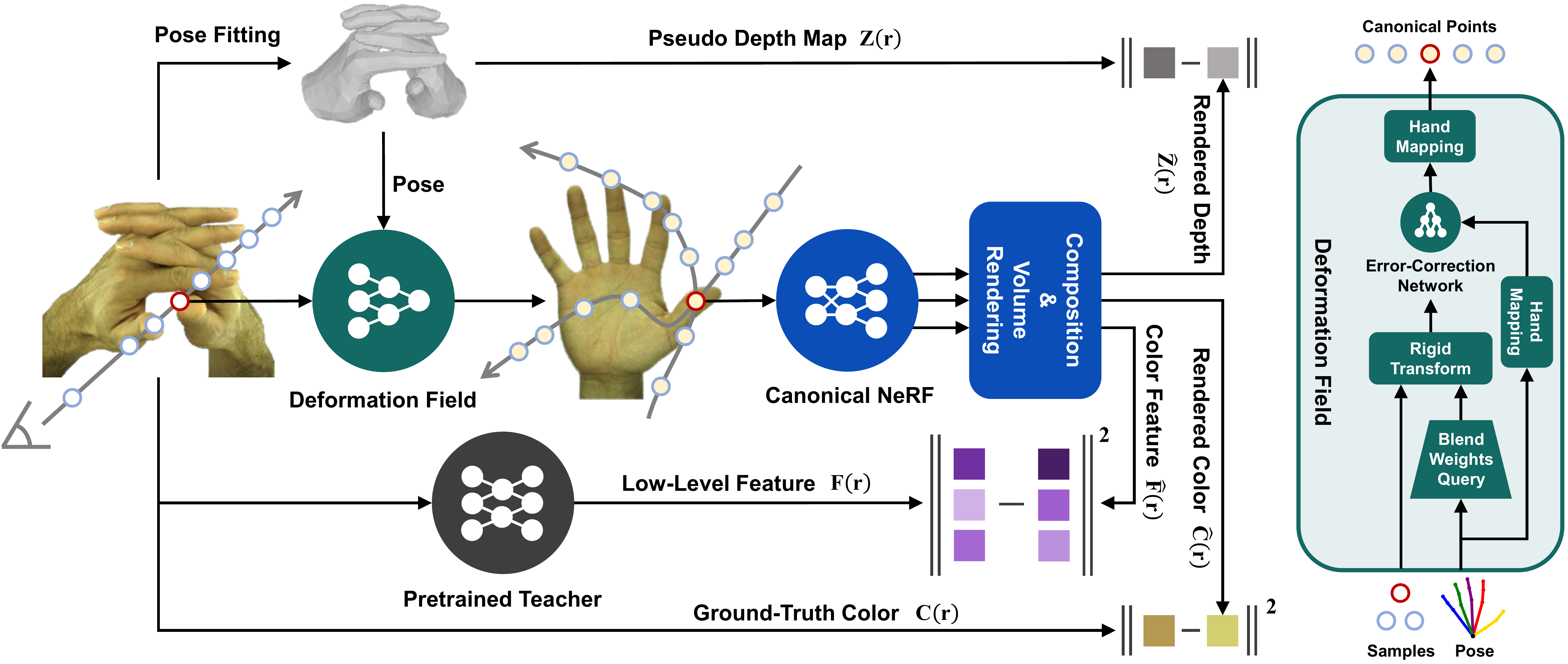}
  \caption{\textbf{Overview of HandNeRF.} A straight observing ray is warped to a canonical space by the deformation field, depending on the different poses of two hands. Colors and densities of the two sets of samples are then produced by the shared NeRF. We establish supervision for the integrated colors, color features and depth values, to help reconstruct fine-grained details of both texture and geometry.}
  \label{fig:pipeline}
\end{figure*}

\vspace{-2mm}
\section{Related Work}
\vspace{-1mm}
\subsection{Neural Radiance Fields (NeRF)}
\vspace{-1mm}
Recent years have witnessed the rapid development of neural implicit representations~\cite{mescheder2019occupancy,deepsdf,nerf} in 3D modeling and image synthesis. Compared with classic discrete counterparts such as meshes, point clouds, and voxels, neural implicit representations model the scene with neural networks, which are spatially continuous and indicate a higher fidelity and flexibility.
As the most popular implicit representation in neural rendering, Neural Radiance Fields (NeRF)~\cite{nerf} has exhibited stunning results in various tasks since its first introduction. The original NeRF overfits on one static scene by design, therefore it cannot model time-varying contents.

Many efforts have been made to adapt NeRF to dynamic scenes.
Some works condition the NeRF with local~\cite{pixelnerf,ibrnet} or global scene representations~\cite{nsff,gao2021dynamic,du2021neural} to implicitly provide generalizability for it.
As the pioneers, D-NeRF~\cite{dnerf} and Nerfies~\cite{nerfies} use an explicit deformation field to bend straight rays passing through varying targets into a common canonical scene, where a conventional NeRF is optimized. Such a pipeline is adopted by many follow-up works~\cite{hypernerf,tretschk2021non}. These methods provide hard constraints by sharing geometry and appearance information across time, while presenting a relatively harder optimization problem.

\subsection{Neural Rendering of Articulated Objects}
The image rendering of animatable articulated objects, \ie, human bodies, hands, \etc, can be regarded as a special case of modeling dynamic scenes. Most early works~\cite{smpl,mano} complete reconstruction using skeleton-based meshes, which generally rely on expensive calibration and massive samples to produce high-quality results.

Neural Body~\cite{neuralbody} signifies a breakthrough in low-cost human rendering by combining NeRF with the mesh-based SMPL model~\cite{smpl}.
Neural Actor~\cite{neuralactor} optimizes the human model in a canonical space along with a volume deformation based on the linear blending skinning (LBS) algorithm of SMPL mesh.
Similar LBS-based pipelines are adopted by a lot of works~\cite{anerf,aninerf,humannerfe,humannerff,neuman,tava}.
Since the LBS deformation cannot handle non-rigid transformation, other strategies have to be introduced for better rendering quality. Most methods~\cite{humannerff,neuman} regress an extra point-wise offset for samples, while some works like Animatable-NeRF~\cite{aninerf} try to jointly optimize NeRF with the LBS weights for deformation. To this end, a forward and a backward skinning field are introduced to save LBS weights for the bidirectional mapping between the posed and canonical shapes. The main limitation here is the poor generalizability of inverse LBS since the weights vary when the pose changes~\cite{unif}.

Another series of methods~\cite{nasa,narf} model the human body with separate parts. They decompose an articulated object into several rigid bones, and then perform per-bone prediction with separated NeRFs. Although being good at maintaining partial rigidity, those methods struggle to merge different parts. They inevitably produce overlap or breakage between bones, and are consequently inferior to the overall modeling approaches in terms of pose generalizability.
As a result, LBS-based methods are still the mainstream practice for human modeling.
Aside from NeRF, some works~\cite{unif} adopt neural implicit surfaces such as the signed distance field (SDF)~\cite{unisurf,neus,volsdf} to better model the geometry of human body.
Those methods can produce relatively smoother surface predictions, but are not good at rendering appearance with high-frequency details, unlike NeRF.

Compared with human body, the neural rendering of human hands still lacks exploration. Recently, LISA~\cite{lisa} is proposed as the first neural implicit model of textured hands. It is focused on the reconstruction of hand geometry using separately-optimized SDF, while the color results are barely satisfactory. Meanwhile, it suffers from similar limitations as faced by the aforementioned SDF-based and per-bone optimizing methods. Moreover, it only supports one single hand and cannot be applied in the interacting-hand scenarios that are common in practice.


\section{Method}
\label{sec:method}


Given a set of multi-view RGB videos capturing a short pose sequence of a single hand or two interacting hands, we propose a novel framework named HandNeRF, which is intended to model the dynamic scene, enabling image rendering of novel hand poses from arbitrary viewing directions.
The overview of HandNeRF is shown in \cref{fig:pipeline}.
We disentangle the pose of both hands using a deformation field and optimize a shared canonical hand with NeRF (\cref{sec:model}). To ensure the correct depth relationship when compositing two hands, we further establish depth supervision for density optimization (\cref{sec:depth}). Moreover, to mine useful cues from RGB images for better texture learning, we propose a feature distillation framework compatible with our efficient sampling strategy (\cref{sec:distill}).
We will elaborate our method in the following subsections.

\subsection{Preliminary: Neural Radiance Fields}
\label{sec:nerf}

We first quickly review the standard NeRF model~\cite{nerf} for a self-contained interpretation. 
Given a 3D coordinate $\mathbf{x}$ and a viewing direction $\mathbf{d}$, NeRF queries the view-dependent emitted color $\mathbf{c}$ and density $\sigma$ of that 3D location using a multi-layer perceptron (MLP).
A pixel color $\hat{\mathbf{C}}(\mathbf{r})$ can then be obtained by integrating the colors of $N$ samples along a ray $\mathbf{r}$ in the viewing direction $\mathbf{d}$ using the differentiable discrete volume rendering function~\cite{nerf}: 

\vspace{-3mm}
\begin{equation}
    \hat{\mathbf{C}}(\mathbf{r})=\sum_{i=1}^N T_i\left(1-\exp \left(-\sigma_i \delta_i\right)\right) \mathbf{c}_i,
\label{eq:volume_rendering}
\vspace{-1mm}
\end{equation}
where $\delta_i$ is the distance between adjacent samples, and $T_i=\exp (-\sum_{j=1}^{i-1} \sigma_j \delta_j)$.
To further obtain the multi-scale representation of a scene, Mip-NeRF~\cite{mipnerf} extends NeRF to represent the samples along each ray as conical frustums, which can be modeled by multivariate Gaussians $(\mathbf{x}, \mathbf{\Sigma})$, with $\mathbf{x}$ as the mean and $\mathbf{\Sigma} \in \mathbb{R}^{3\times 3}$ as the covariance.
Thus, the density and emitted color for a sample can be given by the NeRF MLP: $(\mathbf{x}, \mathbf{\Sigma}, \mathbf{d}) \rightarrow (\mathbf{c}, \sigma)$.

\subsection{Modeling Pose-Driven Interacting Hands}
\label{sec:model}

The conventional NeRF is optimized on a static scene and lacks the ability to model hands with different poses. Therefore, for pose-driven hands modeling, we introduce a pose-conditioned deformation field that warps the observing rays passing through both hands to a shared space, where a static NeRF is established for one canonical hand.

\parsection{Canonical hand representation}
We model the geometry and texture of hands with a neural radiance field in a pose-independent canonical space. Considering the multi-scale distribution of observers in practice, a cone-tracing architecture similar to Mip-NeRF~\cite{mipnerf} is adopted. To be specific, two MLPs denoted by $F_{\mathbf{\Theta}_\sigma}$ and $F_{\mathbf{\Theta}_c}$ output the density $\sigma$ and emitted color $\mathbf{c}$ of the queried 3D sample, respectively:
\vspace{-3mm}
\begin{align}
    \sigma&=
    F_{\mathbf{\Theta}_\sigma}\left(\operatorname{IPE}\left(\mathbf{x}_{can}, \mathbf{\Sigma}\right)\right)=
    F_{\mathbf{\Theta}_\sigma}\left({\mathbf{f}_\sigma}\right), \\
    \mathbf{c}&=
    F_{\mathbf{\Theta}_c}\left(\operatorname{PE}\left(\mathbf{d}\right),\mathbf{f}_\sigma,\ell_c\right),
\label{eq:color}
\vspace{-3mm}
\end{align}
where $\mathbf{x}_{can}$ is the sample coordinate in the canonical space, $\operatorname{PE}(\cdot)$ is the sinusoidal positional encoding in \cite{nerf}, $\operatorname{IPE}(\cdot)$ is the anti-aliased integrated positional encoding proposed by \cite{mipnerf}, and $\ell_c$ is a per-frame latent code to model subtle texture differences between frames. Definitions of other notations are consistent with those in \cref{sec:nerf}.

\parsection{Deformation field}
Given an arbitrary hand pose, the deformation field is intended to learn a mapping from that observation space to a canonical space shared by all posed hands. Without any motion priors, it is an extremely under-constrained problem to model the deformation field as a trainable pose-conditioned coordinate transformation jointly-optimized with NeRF~\cite{dnerf,nsff}.
Therefore, we follow previous works on NeRF for dynamic human body~\cite{aninerf,humannerff,humannerfe,neuman} to leverage the parameterized human priors.
Specifically, to establish a pose-driven deformation field, HandNeRF follows the settings of MANO~\cite{mano} with the 16 hand joints, the pose parameters $\mathbf{p} \in \mathbb{R}^{16\times3}$ (axis angles at each joint), the canonical (mean/rest) pose $\overline{\mathbf{p}}$, and the blend skinning weight $\mathbf{w}_b \in \mathbb{R}^{16}$.
Similar to many classic mesh-based methods, MANO uses linear blend skinning (LBS) to accomplish skeleton-driven deformation for mesh vertices. It models the coordinate transformation between poses as the accumulation of joints' rigid transformations weighted by the blend weight $\mathbf{w}_b$.

HandNeRF employs such skeleton-driven transformation as a strong prior for the deformation field. Given a pose $\mathbf{p}$ and a 3D sample $\mathbf{x}_{ob}$ from the observation space, we obtain the posed MANO mesh and query the nearest mesh facet for $\mathbf{x}_{ob}$. The queried blend weight $\mathbf{w}_b=[w_{b,1},\dots,w_{b,16}]$ is then calculated by barycentric interpolating those of corresponding facet vertices. Thus, a coarse deformation can be expressed by
\vspace{-2mm}
\begin{equation}
    \hat{\mathbf{x}}_{can} = T\left(\mathbf{x}_{ob}, \mathbf{p}\right) = (\sum_{j=1}^{16}{w_{b,j} \mathbf{T}_j}) \mathbf{x}_{ob},
\label{eq:lbs}
\vspace{-2mm}
\end{equation}
where $\mathbf{T}_j \in \operatorname{SE}(3)$ is the observation-to-canonical rigid transformation matrix of each joint.

Due to the inevitable errors caused by the interpolation and the parameterized model itself, we introduce an additional pose-conditioned error-correction network denoted by $F_{\mathbf{\Theta}_e}$ to model the non-linear deformation as a residual term for $\hat{\mathbf{x}}_{can}$.
In this way, the deformation field can capture pose-specific details beyond the mesh estimation while preserving the generalizability of the canonical hand.

To enable the complementation of geometry and texture for left and right hands in textureless or rarely-observed areas during training, we propose a unified modeling of canonical space for both hands. Since the pose parameters and canonical pose of two hands are defined differently in MANO, we introduce a hand mapping module denoted by $\psi(\cdot)$ in practice to align the left hand with the right one. Formally, the deformation field (illustrated in \cref{fig:pipeline}, right) can be expressed by
\vspace{-2mm}
\begin{equation}
    \mathbf{x}_{can} =
    \psi\left(
    \hat{\mathbf{x}}_{can} + 
    F_{\mathbf{\Theta}_e}\left(
    \psi\left(\hat{\mathbf{x}}_{can}\right),
    \psi\left(\mathbf{p}\right)
    \right)
    \right).
\label{eq:displ}
\vspace{-2mm}
\end{equation}
Note that different from previous works~\cite{anerf,aninerf} relying on per-pose latent code to guide the deformation, we use pose representation instead, ensuring robustness to unseen poses.

\parsection{Sampling and composition strategy}
Based on the estimated parameterized hand mesh, it is convenient to obtain the coarse scene bounds of both 3D space and 2D image.
The 2D image bounds serve as a pseudo label of the foreground mask, which guides the pixel (ray) sampling. For a high-resolution training image, we perform ray-tracing on only 1\% of the pixels, mainly focusing on the foreground. Since the target hand covers only a small area of a typical image, such an unbalanced pixel sampling strategy ensures that more importance is attached to the texture of the target hands, and also significantly speeds up the training.

Meanwhile, the 3D scene bounds help to determine the near and far bounds for a camera ray, along which $N$ 3D samples are evenly selected. In order to render two interacting hands while the canonical NeRF only models a single hand, we have to perform object composition before volume rendering.
Instead of introducing an extra composition operator (\eg, density-weighted mean of colors~\cite{giraffe}), we argue that for each pixel, sampling twice within both hands' own bounds is more reasonable for non-transparent targets without clipping.
Specifically, a straight observing ray is warped with two different solutions produced by the deformation field, depending on the corresponding poses of two hands. The colors and densities of the two sets of deformed samples are produced by the shared canonical NeRF, and then re-sorted based on their depth values. Finally, we integrate over all the samples belonging to the same ray using \cref{eq:volume_rendering} and obtain the final pixel color.

\subsection{Depth-Guided Density Optimization}
\label{sec:depth}

The conventional NeRF is susceptible to visual overfitting when given insufficient training views~\cite{dsnerf}. That is, even if the scene geometry (density) fails to be correctly extracted, the rendered images from specific camera views can still be fine. However, these seemingly fine color results occur only on training views and will collapse for novel view synthesis.
This will become a catastrophe in our task with sparse training views. Worse still, our composition strategy for interacting hands will exhibit poor performance without a relatively accurate geometry prediction. Obviously, the rendering quality of complex poses such as interlocking hands relies highly on a correct depth relationship.

To address this issue, we establish 2D depth supervision on the optimization of 3D density.
Recent works~\cite{dsnerf,depthnerf} introduce depth constraints to NeRF by running structure-from-motion (SFM) preprocessing to produce sparse 3D point clouds that function as depth labels.
Unlike those works, we leverage the parameterized hand model estimated in \cref{sec:model} at a lower cost.
Once the posed hand mesh is obtained, the depth of each pixel from a specific view is freely available as a byproduct. We then use it to build a pseudo depth map as the ground truth for the training view.
Meanwhile, the pixel-wise depth estimated by NeRF can be derived with volume rendering. For $N$ samples along a ray $\mathbf{r}$, we denote their depth values as $\{t_1,t_2,\dots,t_N\}$. Then we integrate these values with the same weights as \cref{eq:volume_rendering}:
\vspace{-2mm}
\begin{equation}
    \hat{Z}(\mathbf{r})=\sum_{i=1}^N T_i\left(1-\exp \left(-\sigma_i \delta_i\right)\right) t_i,
\label{eq:depth_render}
\vspace{-2mm}
\end{equation}
where $\hat{Z}(\mathbf{r})$ is the estimated depth value of a specified ray.

Our objective is to minimize the difference between $\hat{Z}(\mathbf{r})$ and the target depth map $Z(\mathbf{r})$.
While SFM-based depth-supervised methods aim at minimizing the KL divergence~\cite{dsnerf} or a Gaussian negative log likelihood term~\cite{depthnerf} on the depth, we deem it more reasonable to regularize the pixel-wise smooth $L_1$ distance in HandNeRF. That is because unlike sparse point clouds with noise, our mesh-based pseudo depth naturally maintains the surface consistency.

\subsection{Neural Feature Distillation}
\label{sec:distill}

In a conventional NeRF pipeline, the multi-view training images are only used for independent pixel-wise supervision. However, with such a vanilla training framework, artifacts and blurs can often be observed in our task for unseen views or poses with sparse training views. Besides, the model is prone to local optimum on some training sequences due to the miniature visible hand in specific views. All these phenomena call for the re-usage of training images to give attention to the spatial context of individual pixel and impose more constraints on color learning.

Unlike image-based extensions~\cite{pixelnerf,ibrnet} for NeRF that directly feed pixel features learned with a jointly-optimized feature extractor into the color fields, we adopt a more efficient and general method --- neural feature distillation.
Our objective is to align the 2D image features, produced by a pre-trained extractor, with the corresponding sample features defined in 3D space. Therefore, contextual cues can be implicitly introduced to the optimization of color field, owing to the receptive field of the feature extractor.

Specifically, we adopt a cross-domain student-teacher paradigm, where features of a 2D teacher network are distilled into a 3D student network. Instead of learning an extra neural feature field as in N3F\cite{n3f}, we make the NeRF output a color feature $\mathbf{f}_c\in{\mathbb{R}^D}$, where $D$ is the number of feature channels. $\mathbf{f}_c$ is derived from the viewing direction $\mathbf{d}$ and the density feature $\mathbf{f}_\sigma$, as an intermediate product of the color field in \cref{eq:color}. Then $\mathbf{f}_c$ of all samples along the sampled ray is integrated using volume rendering (\cref{eq:volume_rendering}) to produce a pixel-wise feature $\hat{\mathbf{F}}(\mathbf{r})$.
As for the 2D teacher network, we choose the self-supervised extractor DINO~\cite{dino} built based on vision transformer. Note that other popular image feature extractors~\cite{mocov3,mae} can also be applied in our framework.
The target image feature is extracted from the second layer of the pre-trained DINO using the publicly available weights, which is meant to focus on texture details rather than high-level semantics. It is then $L_2$-normalized and reduced to $D$ dimensions with PCA before distillation, yielding the target pixel feature $\mathbf{F}(\mathbf{r})$.

\subsection{Training}
\label{sec:train}

\parsection{Loss function}
Following \cite{nerf}, the main loss for optimizing the NeRF network parameters $\mathbf{\Theta}_\sigma$ and $\mathbf{\Theta}_c$ is applied directly between the rendered pixel color $\hat{\mathbf{C}}(\mathbf{r})$ and the ground truth $\mathbf{C}(\mathbf{r})$:
\vspace{-3mm}
\begin{equation}
    \mathcal{L}_{rgb}=
    \sum_{\mathbf{r}}
    \Vert\mathbf{C}(\mathbf{r}) - \hat{\mathbf{C}}(\mathbf{r})\Vert_2^2.
\vspace{-2mm}
\end{equation}

As mentioned in \cref{sec:depth}, we propose an extra constraint on $\mathbf{\Theta}_\sigma$, regularizing the pixel-wise distance between the rendered depth $\hat{Z}(\mathbf{r})$ and the target pseudo depth $Z(\mathbf{r})$:
\vspace{-1mm}
\begin{equation}
    \mathcal{L}_{depth}=
    \sum_{\mathbf{r}}
    \operatorname{SLL}(Z(\mathbf{r}) - \hat{Z}(\mathbf{r})),
\vspace{-2mm}
\end{equation}
where $\operatorname{SLL}(\cdot)$ is the smooth $L_1$ loss.

As interpreted in \cref{sec:distill}, the neural distillation is performed on the color feature $\hat{\mathbf{F}}(\mathbf{r})$ to achieve cross-domain alignment:
\vspace{-3mm}
\begin{equation}
    \mathcal{L}_{dst}=
    \sum_{\mathbf{r}}
    \Vert\mathbf{F}(\mathbf{r}) - \hat{\mathbf{F}}(\mathbf{r})\Vert_2^2.
\vspace{-3mm}
\end{equation}
Note that $\mathcal{L}_{color}$, $\mathcal{L}_{depth}$ and $\mathcal{L}_{dst}$ are also back propagated to update parameters of the deformation field, $\mathbf{\Theta}_e$.

Besides, we add a regularizer for the error-correction term of each sample $\mathbf{x}$ in the deformation field (\cref{eq:displ}), so that the non-linear deformation is minor and does not degrade the generalizability for unseen poses:
\vspace{-1mm}
\begin{equation}
    \mathcal{L}_{dfm}=
    \sum_{\mathbf{x}}
    \Vert
    F_{\mathbf{\Theta}_e}\left(
    \psi\left(\hat{\mathbf{x}}_{can}\right),
    \psi\left(\mathbf{p}\right)\right)
    \Vert_2.
\vspace{-3mm}
\end{equation}

Additionally, to mitigate the semi-transparent geometry and the misty halo around the target hand, we apply the hard surface loss similar to~\cite{lolnerf}, encouraging the weight of each sample in volume rendering to be either $1$ or $0$:
\vspace{-2mm}
\begin{equation}
    \mathcal{L}_{hs}=
    \sum_{\mathbf{x}}
    -\log(e^{-|w_v|}+e^{-|1-w_v|}),
\vspace{-2mm}
\end{equation}
where $w_v=T_i\left(1-\exp \left(-\sigma_i \delta_i\right)\right)$ is the weight in \cref{eq:volume_rendering}.

Moreover, we observe that on some sequences, our model is prone to a local optimum where all pixels on the target hands are converged to the same mean color. We have to impose stronger regularization for samples that are closer to the mean. To this end, a color variance loss is proposed:
\vspace{-1mm}
\begin{equation}
    \mathcal{L}_{cvar}=
    \operatorname{SLL}(
    \operatorname{Var}(\{\mathbf{C}\}) - \operatorname{Var}(\{\hat{\mathbf{C}}\})
    ),
\vspace{-1mm}
\end{equation}
where $\operatorname{Var}(\cdot)$ calculates the biased sample variance.

Overall, the final loss is given by

\vspace{-3mm}
\begin{gather}
\begin{split}
    \mathcal{L}=
    &\mathcal{L}_{rgb}+\lambda_{depth}\mathcal{L}_{depth}+\lambda_{dst}\mathcal{L}_{dst}+\\
    &\lambda_{dfm}\mathcal{L}_{dfm}+\lambda_{hs}\mathcal{L}_{hs}+\lambda_{cvar}\mathcal{L}_{cvar}
    .
\end{split}
\vspace{-2mm}
\end{gather}

\parsection{Pose generalization and adaptation}
Once trained, our model is able to produce full-resolution images for novel poses as well as novel views. Due to our design of fully manipulable pose input, rendering animatable interacting hands is as simple as feeding the desired pose parameters into HandNeRF. The mesh priors and our canonical hand model ensure the generalizability for out-of-distribution poses. Nevertheless, if the training pose sequences are too homogeneous, HandNeRF may still fail to disentangle pose-specific shapes (\eg, the tense muscles) from the canonical geometry, resulting in conspicuous artifacts for novel poses. Fortunately, our framework can be conveniently modified into a fine-tuning pipeline for pose adaptation. Specifically, we disable the feature distillation branch, freeze the parameters of NeRF, and fine-tune the deformation field. Only the depth and the deformation loss are used in this stage. No ground-truth RGB image is needed, as the depth map can be derived directly from pose parameters. After pose adaptation on a few samples, the rendered images will have much fewer artifacts and geometric errors.

\section{Experiments}
\label{sec:exp}

\begin{figure*}[t]
  \centering
  \includegraphics[width=1.0\linewidth]{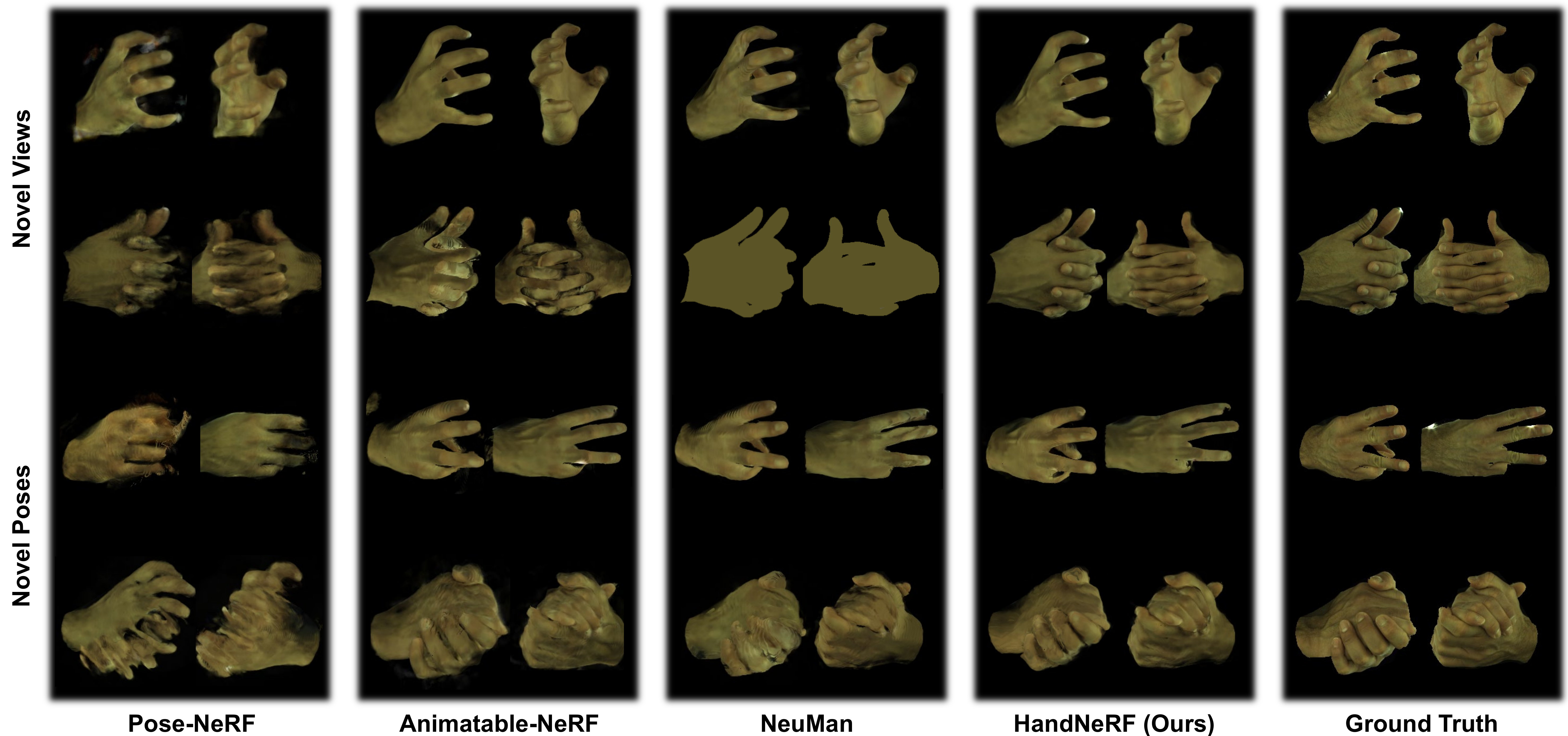}
  \vspace{-6mm}
  \caption{\textbf{Qualitative performance comparison.} We present the results of both novel view rendering (first two rows) and novel pose adaptation (last two rows). All models are trained with 10 different camera views. Pose adaptation results for interacting hands (last row) are produced with models pre-trained on one single hand.}
  \label{fig:results}
\end{figure*}

\subsection{Experimental Settings}

\parsection{Dataset and preprocessing}
HandNeRF is trained on the 30FPS version of Interhand2.6M~\cite{interhand} that contains large-scale multi-view sequences of various hand poses. Each sequence contains images (512 $\times$ 334 px) of a single hand or interacting hands from dozens of views. 18 common views are selected as the test views. Intra-sequence test is to evaluate the novel view synthesis quality, while cross-sequence test is to evaluate the novel pose rendering quality.

\parsection{Baselines}
HandNeRF is the first NeRF model designed for photo-realistic novel view/pose image synthesis of interacting hands, thus no method is available for direct comparison. Therefore, we develop three baselines inspired by works that explore NeRF for human body.
1) Pose-NeRF: we modify Mip-NeRF~\cite{mipnerf} to learn a NeRF conditioned on pose;
2) Ani-NeRF: we adapt \cite{aninerf} to the setup of human hands;
3) NeuMan: we re-implement the ``Human NeRF'' module of \cite{neuman} on the settings of hands while preserving its various training losses.
We do not include LISA~\cite{lisa} because its source code and customized datasets are unavailable for a fair comparison.
Since all the above baselines are for one single articulated object only, we extend them with the proposed composition strategy in HandNeRF to integrate two independent canonical models for both hands.

\parsection{Metrics}
Following previous works, we evaluate the synthesized results with peak signal-to-noise ratio (PSNR), structural similarity index (SSIM), and learned perceptual image patch similarity (LPIPS). To show the effect of our proposed depth supervision, we additionally provide the average $L_1$ error for the rendered depth map (DE), representing the quality of geometric reconstruction to some extent.

\subsection{Comparison Results}
\label{sec:results}

\begin{table*}
  \centering
  \resizebox{1\linewidth}{!}{
    \begin{tabular}{c|cccc|cccc|cccc}
    \hline
          & \multicolumn{4}{c|}{4 views}  & \multicolumn{4}{c|}{7 views}  & \multicolumn{4}{c}{10 views} \\
          & PSNR $\uparrow$ & SSIM $\uparrow$ & LPIPS $\downarrow$ & DE $\downarrow$ & PSNR $\uparrow$ & SSIM $\uparrow$ & LPIPS $\downarrow$ & DE $\downarrow$ & PSNR $\uparrow$ & SSIM $\uparrow$ & LPIPS $\downarrow$ & DE $\downarrow$ \\
    \hline
    \multicolumn{13}{c}{Single hand} \\
    \hline
    Pose-NeRF & 27.0855  & 0.9355  & 0.0921  & 0.1517  & 29.2643  & 0.9301  & 0.0704  & 0.1855  & 29.2126  & 0.9397  & 0.0739  & 0.1910  \\
    Ani-NeRF & 30.2606  & 0.9589  & 0.0704  & 0.1700  & 31.6422  & 0.9632  & 0.0581  & 0.1623  & 31.7784  & 0.9684  & 0.0621  & 0.1582  \\
    NeuMan & 30.3428  & 0.9596  & 0.0691  & 0.1685  & 31.2364  & 0.9623  & 0.0573  & 0.1617  & 31.8419  & 0.9702  & 0.0552  & 0.1507  \\
    Ours  & \textbf{31.0493 } & \textbf{0.9655 } & \textbf{0.0588 } & \textbf{0.1278 } & \textbf{31.8556 } & \textbf{0.9691 } & \textbf{0.0459 } & \textbf{0.1238 } & \textbf{32.7036 } & \textbf{0.9742 } & \textbf{0.0375 } & \textbf{0.1210 } \\
    \hline
    \multicolumn{13}{c}{Interacting hands} \\
    \hline
    Pose-NeRF & 25.0193  & 0.8745  & 0.1873  & 0.2604  & 27.2416  & 0.9014  & 0.1381  & 0.2464  & 27.6461  & 0.9162  & 0.1071  & 0.2312  \\
    Ani-NeRF & 28.0323  & 0.9414  & 0.0865  & 0.2260  & 28.8543  & 0.9440  & 0.0841  & 0.2187  & 29.3577  & 0.9491  & 0.0798  & 0.2118  \\
    NeuMan & $\times$ & $\times$ & $\times$ & $\times$ & $\times$ & $\times$ & $\times$ & $\times$ & $\times$ & $\times$ & $\times$ & $\times$ \\
    Ours  & \textbf{29.0351 } & \textbf{0.9555 } & \textbf{0.0841 } & \textbf{0.1861 } & \textbf{30.0691 } & \textbf{0.9624 } & \textbf{0.0818 } & \textbf{0.1863 } & \textbf{30.7571 } & \textbf{0.9568 } & \textbf{0.0724 } & \textbf{0.1864 } \\
    \hline
    \end{tabular}
  }
  \vspace{-2mm}
  \caption{\textbf{Performance comparison on novel view synthesis.} ``$\times$'' means the model does not converge properly on one or more training sequences. Our method achieves the best rendering quality across all scenes, even only trained with extremely sparse views.}
  \vspace{-2mm}
  \label{tab:novelview}
\end{table*}

\begin{table}
  \centering
  \tabcolsep=1mm
  \resizebox{1\linewidth}{!}{
    \begin{tabular}{l|cccc}
    \hline
    & PSNR $\uparrow$  & SSIM $\uparrow$  & LPIPS $\downarrow$ & DE $\downarrow$ \\
    \hline
    \multicolumn{5}{c}{Single hand $\rightarrow$ Single hand} \\
    \hline
    Pose-NeRF & 23.0118 / ~~~~---~~~~ & ~0.8959 / ~~~~---~~~~ & ~0.1454 / ~~~~---~~~~ & ~0.1985 / ~~~~---~~~~ \\
    Ani-NeRF & ~~~~---~~~~~ / 25.0533 & ~~~~---~~~ / 0.9317 & ~~~~---~~~ / 0.0742 & ~~~~---~~~ / 0.1596 \\
    NeuMan & 25.0254 / 25.8456 & 0.9258 / 0.9324 & 0.0955 / 0.0608 & 0.1605 / 0.1321 \\
    Ours  & \textbf{26.5088 / 27.9717 } & \textbf{0.9345 / 0.9532 } & \textbf{0.0911 / 0.0576} & \textbf{0.1435 / 0.1279} \\
    \hline
    \multicolumn{5}{c}{Single hand $\rightarrow$ Interacting hands} \\
    \hline
    Pose-NeRF & 21.1971 / ~~~~---~~~~ & ~0.8344 / ~~~~---~~~~ & ~0.1959 / ~~~~---~~~~ & ~0.2137 / ~~~~---~~~~ \\
    Ani-NeRF & ~~~~---~~~~~ / 23.9512 & ~~~~---~~~ / 0.9218 & ~~~~---~~~ / 0.0934 & ~~~~---~~~ / 0.1800 \\
    NeuMan & 24.0815 / 24.9451 & 0.9104 / 0.9283 & 0.1203 / 0.0951 & 0.1766 / 0.1626 \\
    Ours  & \textbf{25.4666 / 26.5207} & \textbf{0.9180 / 0.9348} & \textbf{0.1162 / 0.0897} & \textbf{0.1652 / 0.1601} \\
    \hline
    \multicolumn{5}{c}{Interacting hands $\rightarrow$ Interacting hands} \\
    \hline
    Pose-NeRF & 19.8561 / ~~~~---~~~~ & ~0.8468 / ~~~~---~~~~ & ~0.2321 / ~~~~---~~~~ & ~0.1954 / ~~~~---~~~~ \\
    Ani-NeRF & ~~~~---~~~~~ / 23.0223 & ~~~~---~~~ / 0.8928 & ~~~~---~~~ / 0.1465 & ~~~~---~~~ / 0.2221 \\
    NeuMan & $\times$~~~~ / ~~~~$\times$ & $\times$~~~~ / ~~~~$\times$ & $\times$~~~~ / ~~~~$\times$ & $\times$~~~~ / ~~~~$\times$ \\
    Ours  & \textbf{23.6411 / 24.8599} & \textbf{0.8945 / 0.9152} & \textbf{0.1315 / 0.0858} & \textbf{0.1835 / 0.1802} \\
    \hline
    \end{tabular}
  }
  \vspace{-2mm}
  \caption{\textbf{Performance (generalization / adaptation) comparison on novel pose synthesis.} ``---'' means the method is inapplicable for that setup. ``$\times$'' means the model does not converge properly on previous training. The hand types on both sides of ``$\rightarrow$'' indicate the NeRF training samples and novel pose samples.
  Since Ani-NeRF~\cite{aninerf} cannot directly generalize to unseen poses, we report its pose adaptation performance after re-training with blend weight consistency. Due to the local optimum results of NeuMan~\cite{neuman} on interacting hands, we exclude it in those comparisons.}
  \label{tab:novelpose}
\end{table}

\cref{tab:novelview} and \ref{tab:novelpose} summarize the performance of HandNeRF and the baselines.
Qualitative results are exhibited in \cref{fig:results}.

\parsection{Novel view synthesis}
We train the model on a single sequence with 4, 7, or 10 views to show the effect of view quantity. As presented in \cref{tab:novelview}, our method outperforms all the baselines across all metrics.
Notably, training a model only with interacting hands samples is a non-trivial task, since it involves self-occlusion, incompleteness of visible texture, and subtle contacts during interaction.
Therefore, the superiority of our proposed unified modeling can be evidently observed from the results.
Even trained with extremely sparse views, HandNeRF can still achieve 29dB for interacting hands in terms of PSNR.
For comparison, NeuMan~\cite{neuman} fails to converge properly on interacting hands, rendering mask-like textureless images. Similar issue also arises during the training of HandNeRF, but we manage to resolve it with the proposed color variance loss.
Besides, we can observe some semi-transparent mist floating around the rendered hand in some methods' results (\cref{fig:results}), which proves the effectiveness of the proposed depth-guided density optimization in HandNeRF.

\parsection{Novel pose synthesis}
We first directly test the learned model on an unseen sequence. Then we apply pose adaptation for those novel poses and re-test the performance.
Three different tasks are included, where the learned model of single hand or interacting hands is generalized or adapted to both hand types. Obviously, it is most challenging to render novel poses for interacting hands when the learned model is also trained on interacting hands.
As shown in \cref{tab:novelpose}, HandNeRF gives the best performance in both pose generalization and further adaptation for all tasks.
Note that the training texture details (usually different from test hands) are preserved in novel pose synthesis, and we do not further optimize color in pose adaptation. Therefore, it is not surprising that pixel-wise metrics like PSNR drop significantly for novel poses. As a perceptual metric, LPIPS is considered to be more meaningful here.

\subsection{Ablation Study}
\label{sec:ablation}

\begin{table}
  \centering
  \tabcolsep=4mm
  \resizebox{1\linewidth}{!}{
  \begin{tabular}{l|cccc}
    \hline
     & PSNR $\uparrow$ & SSIM $\uparrow$ & LPIPS$ \downarrow$ & DE $\downarrow$\\
    \hline
    w/o $\mathcal{L}_{depth}$ & 30.1057  & 0.9528  & 0.0755  & 0.2106  \\
    w/ GNLL & 30.4304  & 0.9552  & 0.0845  & 0.1852  \\
    Ours  & \textbf{30.9256} & \textbf{0.9570} & \textbf{0.0700} & \textbf{0.1840} \\
    \hline
    w/o distillation & 32.8421  & 0.9720  & 0.0488  & 0.1361  \\
    random distillation & 32.7892  & 0.9712  & 0.0506  & 0.1361  \\
    w/ CNNRenderer & 31.6366  & 0.9703  & 0.0493  & 0.1362  \\
    w/ TransRenderer & 31.3229  & 0.9680  & 0.0479  & 0.1364  \\
    Ours  & \textbf{33.0204} & \textbf{0.9737} & \textbf{0.0475 } & \textbf{0.1360} \\
    \hline
  \end{tabular}
  }
  \vspace{-2mm}
  \caption{\textbf{Ablation study.} Experiments about depth supervision are performed on interacting hands, while the results of neural distillation and neural renderer are produced on one single hand.}
  \label{tab:ablation}
\end{table}%

\begin{figure}[t]
  \centering
  \vspace{-2mm}
  \includegraphics[width=1.0\linewidth]{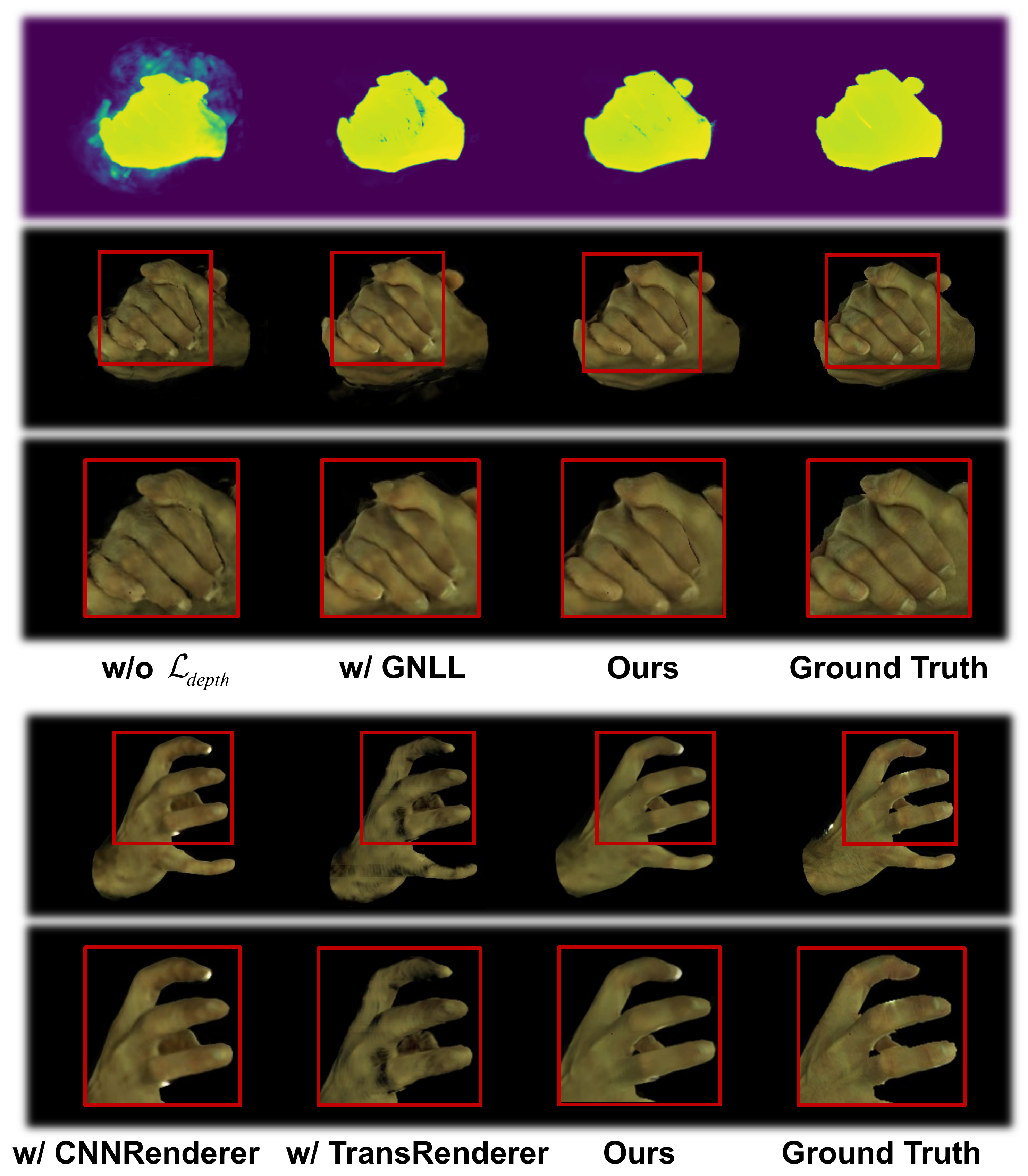}
  \vspace{-7mm}
  \caption{\textbf{Visualization of ablation study.} We exhibit rendering results and zoomed-in details of ablations for depth supervision (upper) and neural renderer (lower). Additionally, the rendered depth maps are shown in the first row.}
  \label{fig:ablation}
\end{figure}

We conduct ablative experiments to validate the effectiveness of two essential components in our HandNeRF: the depth-guided density optimization and the neural feature distillation. The results are listed in ~\cref{tab:ablation}. We also provide more details of the synthesized images in \cref{fig:ablation}.

\parsection{Depth supervision}
We ablate our depth supervision and compare it with GNLL (Gaussian negative log likelihood)~\cite{depthnerf}.
We observe dramatic performance degradation and blurred interfacial areas without depth guidance. Although overfitting to training views, the model fails to infer correct depth when rendering from novel views, let alone compositing both hands for unseen poses. This can be further proved by its noisy depth map.
Besides, as mentioned in \cref{sec:depth}, GNLL is more suitable for depth values produced by noisy point clouds. When applied to our mesh-based depth map, it leads to artifacts near hand geometry.

\parsection{Neural distillation}
We replace the image features produced by our pre-trained teacher with the same-shaped random vectors. The results in \cref{tab:ablation} show that HandNeRF actually exploits the texture information from the low-level features. The performance improvement does not come from the effect of regularization.

\parsection{Neural renderer}
To get better rendering results, some works~\cite{giraffe,headnerf} propose a neural renderer alongside the conventional volume rendering.
Technically, they increase the number of channels of emitted color to model the more expressive color features with NeRF, which are integrated using volume rendering to produce a feature map. Then 2D neural networks are adopted to render the final RGB image.
However, due to our efficient ray sampling strategy (\cref{sec:model}), only a few sampled pixels are available during training, resulting in an incomplete feature map.
Since full-resolution ray-tracing has an unacceptable overhead, a compromised solution is to produce a much smaller feature map and perform upsampling in the neural renderer, at a cost of NeRF's expressive power, especially for high-frequency details on small targets like hands.
To compare with those neural renderers, we follow \cite{headnerf} to feed a low-resolution 2D feature map into a modified version of its neural rendering module composed of CNN and upsampling layers. We also develop a TransRenderer that uses the transformer for pixel-wise neighborhood attention. It can be observed in \cref{fig:ablation} that the CNNRenderer produces a visually smoother image but degrades the quantitative results, while the TransRenderer tends to fuse hand skin with background noise.

\section{Conclusion}

In this paper, we propose HandNeRF, a novel framework that reconstructs photo-realistic appearance and geometry of single or interacting hands with pose-deformable neural radiance fields.
By developing several elaborate strategies including depth-guided density optimization and neural feature distillation, our method can effectively handle non-trivial challenges in complex hand interactions (\eg, self-occlusion and invisible texture). We thereby enable the rendering of high-fidelity images and videos for gesture animation from arbitrary views.
Comprehensive experiments on the large-scale InterHand2.6M dataset demonstrate the superiority of our approach.

~\\
\textbf{Acknowledgements.}
This work was supported in part by the National Natural Science Foundation of China under Contract U20A20183 and 61836011, and in part by the Fundamental Research Funds for the Central Universities under contract WK3490000007. It was also supported by the GPU cluster built by MCC Lab of Information Science and Technology Institution and the Supercomputing Center of the USTC.

{\small
\bibliographystyle{ieee_fullname}
\bibliography{egbib}
}

\end{document}